\documentclass[conference]{IEEEtran}
\usepackage{cite}
\usepackage{rotating}
\usepackage{graphicx}
\usepackage{rotating}
\usepackage{amsmath}
\usepackage{algorithmic}
\usepackage{color}
\usepackage{xspace}
\usepackage{array}
\ifCLASSOPTIONcompsoc
  \usepackage[caption=false,font=normalsize,labelfont=sf,textfont=sf]{subfig}
\else
  \usepackage[caption=false,font=footnotesize]{subfig}
\fi
\usepackage{stfloats}
\usepackage{url}

\usepackage{flushend}

\newcommand{\comment}[1]{}

\usepackage[dvipsnames]{xcolor}

\begin{document}
\title{Toward Certification of Machine-Learning Systems for Low Criticality Airborne Applications}

\author{\IEEEauthorblockN{Konstantin Dmitriev}
\IEEEauthorblockA{Technical University of Munich -- Garching, Germany\\
Email: konstantin.dmitriev@tum.de}
\and
\IEEEauthorblockN{Johann Schumann}
\IEEEauthorblockA{KBR, NASA Ames Research Center, Moffett Field, CA\\
Email: johann.m.schumann@nasa.gov}
\and
\IEEEauthorblockN{Florian Holzapfel}
\IEEEauthorblockA{\centerline{Technical University of Munich -- Garching, Germany}\\
Email: florian.holzapfel@tum.de}
}

\maketitle

\graphicspath{{../figures/}}

\makeatletter
\def\ps@IEEEtitlepagestyle{%
\def\@oddfoot{\mycopyrightnotice}%
\def\@evenfoot{}%
}
\def\mycopyrightnotice{%
{\footnotesize DOI: 10.1109/DASC52595.2021.9594467 \copyright~2021 IEEE. Personal use of this material is permitted.
\hfill} % Revise this line accordingly!
\gdef\mycopyrightnotice{}
}

\begin{abstract}
The exceptional progress in the field of machine learning (ML) in recent years has attracted a lot of interest in using this technology in aviation. Possible airborne applications of ML include safety-critical functions, which must be developed in compliance with rigorous certification standards of the aviation industry.
Current certification standards for the aviation industry were developed prior to the ML renaissance without taking specifics of ML technology into account. There are some fundamental incompatibilities between traditional design assurance approaches and certain aspects of ML-based systems.
In this paper, we analyze the current airborne certification standards and show that all objectives of the standards can be achieved for a low-criticality ML-based system if certain assumptions about ML development workflow are applied.
\end{abstract}


\section{Introduction}
\label{sec:intro}
The European Union Aviation Safety Agency (EASA) has recently published a roadmap \cite{easaroadmap2020} envisioning an increasing use of Artificial Intelligence (AI) and, in particular, machine learning systems in commercial aviation in the near future. Major application areas include autonomous flight, preventive maintenance, and air traffic management.
 
However, some important aspects of machine learning principles are incompatible with the requirements of aviation industry standards for safety critical systems, such as granular traceability between requirements and design, and appropriate metrics for structural coverage analysis. The international technical committees EUROCAE WG-114 and SAE G-34 were recently established to develop new industry standards to support development and certification of aeronautical systems leveraging machine learning technology. However, the formal consensus-based process for industry standard creation may take many years, as was the case for the foundational airborne standards DO-178C \cite{RTCA.178} and ARP-4754A \cite{SAE.4754}, %whereas ML technology could already be very useful for airborne systems today.
whereas there is an increasing need for ML technology for airborne systems today.
 
This paper provides an analysis of certifiability of low-criticality (ARP-4754A Design Assurance Level~D) ML-based systems based on the existing industry standards. The objectives and methods of the industry standards are analyzed for applicability and satisfiability in the context of the proposed development workflow for a machine learning component. This analysis is intended to facilitate immediate deployment of machine learning technology in low criticality airborne systems based on the existing regulatory framework without a need for new regulations.

The rest of the paper is structured as follows:
in Section~\ref{sec:relwork} we provide an overview of ML technology and discuss related work.
Section~\ref{sec:DOML} includes the analysis of DO-178C standards with respect to ML technology. We first define the relevant notions  and assumptions about the ML development workflow. Then we review in detail the objectives of the standards focusing on software of low-criticality (DO-178C Level~D).
Section~\ref{sec:OTHER} discusses other relevant aerospace standards and their applicability to machine-learning based systems.
Finally, Section~\ref{sec:concl} summarizes the work and concludes with suggestions for future work.

\section{ML Overview and Related work}
\label{sec:relwork}
In recent years, AI has gained tremendously in popularity and application areas. Most traction was obtained by the use of deep neural networks (DNN) for image analysis, object detection and classification, and natural speech processing.
Neural networks are a subclass of machine learning algorithms, which themselves are a subclass of AI. Therefore, the notion of AI, ML, and DNN are often used interchangeably. In this paper, we focus on DNNs.

In the area of (deep) neural networks, we usually distinguish between the learning phase, where the (D)NN is presented with training data, and the inference phase, where the trained network receives inputs (e.g., from a camera or other sensors) and provides an estimate of the desired output.
Inside the neural network, the information is typically
stored as vectors of numbers, called weight parameters, or  simply {\em weights}. These weights are obtained during the execution of a {\em training algorithm}. Here, usually large amounts of training data are used to set the weights in such a way that the DNN provides best-possible estimates on test data.

There exist a large number of different learning regimes
(e.g., unsupervised or supervised learning, reinforcement learning) and numerous learning algorithms. Also, the neural network can be trained during system development, and the pre-trained network with frozen weight parameters can be used for inference during operation. In contrast, online learning (or adaptation) adjusts the network weights during operation. Thus, the system can learn and adapt toward novel situations while in operation.

For our study, we restrict ourselves to the most popular type of neural networks: a deep multilayer convolutional neural network, which is being trained during development and not retrained during operation. Such a network is defined by its architecture and the values of the weight and bias parameters. Their values are the result of the learning process, using a given set of training data and a given learning algorithm.

Typical examples (in the aerospace domain) for such a network include:
\begin{itemize}
\item
object detection and classification, based upon camera inputs,
for example, the detection of runway debris \cite{runway-debris-detection-2018} and objects \cite{runway-object-detection-2020},
of airport signs, or for the avoidance of runway incursions \cite{CNN-runway-incursions-2019}.
\item NN-based control. Flight-control software can use neural networks for inner-loop control (e.g., Intelligent Flight Control System \cite{IFCS2007}), or for efficient calculation of aircraft dynamics \cite{YuYaoLiu2018}. Other examples include the use of neural networks for Taxiing \cite{taxinet2020}, or vision-based takeoff or landing \cite{vision-based-landing-2019}.
\item
Trajectory prediction. NNs can be used for 4D trajectory prediction \cite{easaL1concept2021} or for on-board collision avoidance, e.g., ACAS~Xu \cite{ACAS-XU-2018}.
\item
Pilot advisory systems \cite{easaL1concept2021}.
\item
Fault detection, diagnostics, and prognostics. Here, NNs are used to recognize system faults, to reason about diagnoses, or to support component or system prognostics \cite{HEO2018470,DNN-diagnostics-2019}.
\item
Natural language processing, e.g., to recognize ATC utterances \cite{aerospace8030065,lowry2019}.
%\item
%\TODO{examples from EASA, RTW architecture(?)}
\end{itemize}

The operation of a neural network (or, more general, an ML component) on-board an aircraft can have, depending on the application,
an impact on safety and performance. Developers of a safety-critical ML-based system have to address numerous issues, ranging from suitable architecture selection, trust and confidence, to testing and explainability \cite{Schumann2010}. Ultimately, any airborne safety-critical system must be developed in compliance with stringent certifications standards, which were released prior to the wide adoption of ML and do not address specific aspects of ML technology.

The technical report \cite{eurocae2021SoC} published by  EUROCAE and SAE joint committee in 2021 identifies gaps in the existing certification standards with respect to ML technology and discusses potential approaches to address the gaps, but without detailed information on specific standards' objectives. Another recent work \cite{delseny2021MLinCertifiedSys} includes a detailed analysis of ML-based system certification challenges without proposing solutions. 

There are multiple ongoing and completed research projects that tackle different problematic aspects of ML design assurance, such as verification and validation \cite{ZHANG2020,Borgetal2018}, traceability \cite{aravantinos2019traceability}, or coverage \cite{sun2019nnCoverage,cheng2018quantitativeCov}. However, novel methods are often in their infancy and there is no systematic summary showing how they will address all incompatibilities in existing standards.

% \Johann{I JUST PUT THAT IN -- probably should be merged with the two paragraphs below. What do you think?}

\section{DO-178C and Machine Learning}
\label{sec:DOML}
DO-178C \cite{RTCA.178} is a foundational standard for aviation software development. It specifies the \textit{objectives} to be satisfied by software developers and provides detailed guidelines for \textit{processes} and \textit{activities} to be implemented, as well as the description of \textit{artifacts} (data) to be produced for demonstrating a necessary \textit{level} of software \textit{assurance}. DO-178C defines the following key terms which we frequently use in this paper:
\begin{itemize}
    \item \textit{Objective} --  requirements that should be met to demonstrate compliance with DO-178C.
    \item \textit{Activity} -- task that provides a means of meeting the objectives.
    \item \textit{Process} -- a collection of activities performed during the software life cycle to produce a definable output or product.
    \item \textit{Design Assurance Level (DAL, Software Level)} -- the designation that is assigned to a component (e.g., software item) as determined by the system safety assessment process. The software level establishes the rigor necessary to demonstrate compliance with DO-178C.
    \item \textit{Assurance} -- the planned and systematic actions necessary to provide adequate confidence and evidence that a product or process satisfies given requirements.
\end{itemize}
DO-178C defines \emph{software levels} from A to E, which correspond to the categories of a software failure condition from "Catastrophic" to "No Safety Effect", respectively. The applicable objectives and independence requirements vary depending on the level. 

In this section we analyze the satisfiability of the DO-178C objectives for the proposed machine learning development workflow. We focus on the Level~D software, for which a limited subset of objectives applies, and show that Level~D objectives can be satisfied in spite of major incompatibilities between ML-based and traditional software development approaches.
 
 \subsection{Proposed ML development workflow}
 In this study, we propose the following assumptions about the machine learning development workflow:
 \begin{enumerate}
    \item An ML-based system typically includes both ML-based components and traditional software components (operating system, hardware support drivers, etc. -- further referred to as 'traditional' components). This analysis is focused on ML components only. %The operation of a neural network (or, more general, an ML component) on-board of an aircraft can have, depending on the application, an impact on safety and performance.
    For traditional non-ML software components, DO-178C applies without modification.
    \item We do not consider adaptive ML systems, which can perform training while the system is in operation and can change or evolve over time. Such ML techniques, for example, based upon reinforcement learning \cite{HKvK2020} add another layer of substantial complexity.
    \item In the analysis below, we consider a deep neural network as a representative example of an ML-based system to demonstrate the key properties of ML-based systems.
    \item We propose to implement the ML learning (training) at system level, and provide a trained ML model to the software life cycle for implementation. A typical algorithm for ML model (e.g., a DNN) is composed of simple arithmetic operations and therefore can directly serve as low-level software requirements, from which source code can be implemented. This assumption is inspired by the MB Example 5 provided in DO-331 \cite{RTCA.331}, which represents a common industry practice of a model-based development approach. A neural network, however, may require supplemental low-level requirements in traditional format to enable translation to source code, for example, textual specification of a mathematical library for matrix operation algorithms.
%\JOHANN{Explain acronym MB?}
 \end{enumerate}
 
\subsection{Development processes}
DO-178C development processes produce software requirements, design, source code, executable object code, and associated trace data. There are multiple known industry practices for the software development workflow, which can be used to satisfy the objectives of DO-178C development processes, such as waterfall, agile, model-based design approach, or a combination thereof \cite{dmitriev2020leanMBD}. The sequence and relationship of the development activities can vary depending on the selected development workflow, but regardless of the  workflow, strong bi-directional traceability must be established and maintained between requirements on all levels, source code, and test data (discussed in DO-178C, Sections 5.5 and 6.5).

\textbf{Traceability Issue.} Traditional software is implemented by a programmer as human-readable source code, which can be traced to certain requirements it implements. In contrast, the functional behavior of an ML model is mainly specified by a bulk of parameters (e.g., NN weights), which are automatically adjusted by a learning algorithm during ML model training. It is assumed to be practically impossible to map and trace values of these auto-tuned parameters to specific functions implemented by the ML model, because an ML model is generally not directly comprehensible by humans. As a result, the DO-178C traceability objectives are not achievable for an ML model. This {\em traceability issue\/} is one aspect of the more general ML {\em explainability challenge\/} \cite{delseny2021MLinCertifiedSys}.

However, since an ML model is positioned as representing only low-level software requirements in the proposed workflow, \textit{the traceability issue has no effect on Level~D software}, because objectives for low-level requirements do not apply for Level~D software. Table~\ref{table_A2} includes details of analysis for each software development objective applicable to Level~D.

\begin{table*}[htb]
\caption{Software Development Objectives}
\label{table_A2}
\footnotesize
\begin{tabular}{|r|p{0.205\textwidth}|c|c|c|c|c|p{0.55\textwidth}|}
\hline
\multicolumn{2}{|c|}{DO-178C / DO-331 Objectives}&\multicolumn{5}{c|}{Software Levels}&\multicolumn{1}{c|}{Analysis}\\
\multicolumn{2}{|c|}{}&A&B&C&D&E&\\
\hline\hline
1&High-level requirements (HLR) are developed.
&x&x&x&x&&
In the proposed ML workflow, a trained NN is provided from system level to software life cycle as \textit{Design Model}, from which source code can be directly developed. In this case, as discussed in DO-331 MB.6.1.3, "the guidance related to high-level requirements should be applied to the requirements from which the model is developed." This objective can be achieved by developing  system requirements specifying functional, operational, performance, and other applicable characteristics in traditional (e.g., textual) form using the applicable DO-178C guidelines for high-level requirements.\\
\hline
2&Derived high-level requirements
are defined and provided to the system
processes, including the system safety
assessment.&x&x&x&x&&
In the same manner as for the previous objectives, to achieve this objective, DO-178C guidelines for derived high-level requirements can be applied to system level requirements (from which then ML model is developed) expressed in traditional form.\\
\hline
3&Software architecture is developed.&x&x&x&x&&
The software architecture defines the software components and their interfaces to enable appropriate grouping of the software functions. Given that a typical ML component (NN) is composed of simple sequential arithmetic operations, we assume that there is no need for an architectural breakdown of the ML function and it can be arranged as a single ML component. However, this objective still applies to traditional software components, which integrate with ML component and can be achieved using the applicable DO-178C guidelines for software architecture.\\ 
\hline
4&Low-level Requirements are developed.&x&x&x&&&
This objective does not apply to Level~D and is excluded from the analysis.\\
\hline
5&Derived low-level requirements are defined and provided to the system
processes, including the system safety
assessment.&x&x&x&&&
This objective does not apply to Level~D and is excluded from the analysis.\\
\hline
6&Source Code is developed&x&x&x&&&
This objective does not apply to Level~D and is excluded from the analysis.\\
\hline
7&Executable Object Code and Parameter Data Item (PDI) Files, if any, are produced
and loaded in the target computer.&x&x&x&x&&
Activities for production and loading of executable object code and PDI files are not different between traditional software and ML components, DO-178C guidelines can be directly applied to achieve this objective.\\
\hline
\end{tabular}
\end{table*}

\subsection{Verification Processes}
The DO-178C verification process includes 43 objectives listed in the Tables~A-3 to A-7 of DO-178C. In the proposed development workflow, many objectives can be achieved using tradition methods such as requirements-based testing, reviews, and analysis of traditional artifacts (e.g., textual requirements, source code, requirements-based test cases, and procedures). The most problematic verification objectives with respect to ML technology relate to coverage assessment and verification of ML model artifacts as discussed hereafter.

\textbf{Coverage issue.} DO-178C requires assessment of (1) requirements coverage by tests and (2) coverage of source code structure by tests. The first step is intended to assess how well the created tests exercise the intended behavior specified by  software requirements (e.g., to confirm that tests exist for each requirement). This is typically performed by review of tests and associated requirements. The second step is intended to determine elements of the source code structure that are not covered by requirements-based tests, which can indicate unintended functionality in the source code or shortcomings in requirements or tests.
\begin{itemize}
    \item In the proposed workflow, an ML model represents low-level software requirements and therefore it must be evaluated how well tests cover the requirements expressed by the ML model. However, as discussed above, an ML model generally cannot be traced to requirements or tests because it is not directly comprehensible by humans ({\em explainability challenge\/}). Therefore, it becomes impossible to demonstrate test coverage of requirements expressed by the ML model, and the corresponding DO-178C coverage objective is not achievable.
    \item The DO-178C metrics for structural coverage include statement, decision, and modified condition/decision coverage (MC/DC). These metrics are efficient for the control flow of traditional algorithms but are not representative for a typical ML model implementation, which has a very simple control flow while the functional behavior is embedded in the data arrays (in the NN domain referred to as weights). As discussed in \cite{DeepXplore}, \textit{"a single randomly picked test input was able to achieve 100\% code coverage"}. Consequently, even though the DO-178C structural coverage objectives are achievable, they are generally not representative for the ML source code. Additional activities have to be performed to detect unintended functionality in the source code and achieve a necessary level of confidence in correctness and completeness of requirements-based tests.
\end{itemize}

\textbf{ML model verification issue.} As long as an ML model represents low-level requirements (per the proposed workflow), the verification objectives of Table~A-4 of DO-178C apply. As discussed above, an ML model is generally not 
directly comprehensible by humans,
% (\textit{explainability challenge}), 
this impacts most of the applicable verification objectives. For example, verification of accuracy, consistency, and traceability is not feasible for non-comprehensible ML models with traditional human review methods. However, ML model testing with test data sets can be claimed to assist with some verification objectives, similar to the simulation approach for model-based systems addressed in DO-331. This is discussed in detail in Section~\ref{sec:OTHER}.
 
Tables~\ref{table_A3} and \ref{table_A6} include further analysis of the Level~D verification objectives listed in Tables~A-3 and A-6 of DO-178C.

Tables~A-4 (Software Design Verification), A-5 (Code Verification), and A-7 (Verification of Verification Results) of DO-178C include only three objectives applicable to Level~D:
\begin{itemize}
    \item "Software partitioning integrity is confirmed". Partitioning techniques are used to ensure isolation between software components of different levels. We focus on Level~D software in this study only, and therefore this objective does not apply and is excluded from the analysis.
    \item "Parameter Data Item File is correct and complete". PDI files are intended to influence the behavior of software without modifying its executable object code. Examples include configuration tables and databases. Even though PDI may seem relevant for capturing ML parameters (e.g., NN weights), this will require one to verify "all behavior of the Executable Object Code resulting from the contents of the PDI File" (DO-178C, 6.6). Such a verification task is practical for simple parameter sets of limited size but is deemed impossible for the large non-comprehensible ML models (another fallout from the \textit{explainability challenge}). Hence, use of PDI is not envisioned for ML models and this objective is excluded from the analysis.
    \item "Test coverage of high-level requirements is achieved". We assume that traditional requirements-based testing will be used for high-level requirements testing (see Objectives~1 and 2 in Table~\ref{table_A6}), consequently the conventional analysis methods (DO-178C, 6.6.4.1) can be used to achieve this objective.
\end{itemize}

The important outcome of the objectives' analysis is that neither traceability nor ML verification issues prevent the fulfillment of the Level~D objectives if the proposed ML  workflow is followed. 

\subsection{Other Processes}
In addition to development and verification processes, DO-178C specifies the supporting processes: planning, configuration management, quality assurance and certification liaisons processes. The objectives of these supporting processes are mainly independent of the technology used for software development and, as discussed hereafter, can be achieved for ML-based system of any assurance level.

\subsubsection{Planning Process}
The DO-178 planning process produces plans and standards that guide the overall software life cycle. Planning process objectives are summarized in Table~A-1 of DO-178C and include definitions of the software life cycle, its processes and activities, selection of the life cycle environment, addressing additional considerations, and review and coordination of the developed plans. The planning process is accomplished by development and review of plans and standards, which should include all aspects specified in the process objectives; there is no difference between a traditional and an ML-based software life cycle. ML-specific considerations, which are not addressed in DO-178C (such as traceability and coverage issues discussed above) have to be covered by the applicant in the life cycle plans as additional considerations (4.1.d of DO-178C).

\subsubsection{Configuration Management Process}
The configuration management process is implemented in conjunction with all other life cycle processes to ensure a "defined and controlled configuration throughout the software life cycle" (DO-178C). It includes data identification, baseline establishment, change control, and data archiving activities. DO-178C configuration management objectives are technology-agnostic and can be applied to data of any nature. Therefore, all objectives can be achieved for ML data using traditional methods. Special attention may need to be taken for configuration management of large data sets which are not typical in traditional software development.

\subsubsection{Quality Assurance and Certification Liaisons Processes}
These are administrative processes intended to provide confidence that the software life cycle is implemented in compliance with  DO-178C and in coordination with certification authorities. Objectives of these processes are generally abstracted from the technologies used for development and can be achieved for ML-based software in the same manner as for traditional software.

\begin{table*}[htb]
\caption{Software Requirements Verification Objectives}
\label{table_A3}
\footnotesize
\begin{tabular}{|r|>{\raggedright}p{0.2\textwidth}|c|c|c|c|c|p{0.55\textwidth}|}
\hline
\multicolumn{2}{|c|}{DO-178C / DO-331 Objectives}&\multicolumn{5}{c|}{Software Levels}&\multicolumn{1}{c|}{Analysis}\\
\multicolumn{2}{|c|}{}&A&B&C&D&E&\\
\hline\hline
1&High-level requirements comply with system requirements.
&i&i&x&x&&
This objective is implicitly satisfied in the proposed ML workflow, because requirements, from which the ML model is developed, represent both system-level and high-level software requirements (see Item 1 under the Table MB.A-3 of DO-331).\\
\hline
2&High-level requirements are accurate and consistent.
&i&i&x&x&&
High-level requirements are expressed in traditional form (e.g., textual) in the proposed ML workflow and can be verified using standard review and analysis methods.\\
\hline
3&High-level requirements are compatible with target computer.
&x&x&&&&
This objective does not apply to Level~D and is excluded from the analysis.\\
\hline
4&High-level requirements are verifiable.
&x&x&x&&&
This objective does not apply to Level~D and is excluded from the analysis.\\
\hline
5&High-level requirements conform to standards.
&x&x&x&&&
This objective does not apply to Level~D and is excluded from the analysis.\\
\hline
6&High-level requirements are traceable to system requirements.
&x&x&x&x&&
Same as the Objective~1, this objective is implicitly satisfied for the requirements from which ML model is developed.\\
\hline
7&Algorithms are accurate.
&i&i&x&&&
This objective does not apply to Level~D and is excluded from the analysis.\\
\hline
\end{tabular}
\end{table*}

\begin{table*}[htb]
\caption{Testing Objectives}
\label{table_A6}
\footnotesize
\begin{tabular}{|r|>{\raggedright}p{0.2\textwidth}|c|c|c|c|c|p{0.55\textwidth}|}
\hline
\multicolumn{2}{|c|}{DO-178C / DO-331 Objectives}&\multicolumn{5}{c|}{Software Levels}&\multicolumn{1}{c|}{Analysis}\\
\multicolumn{2}{|c|}{}&A&B&C&D&E&\\
\hline\hline
1&Executable object code complies with high-level requirements.
&x&x&x&x&&
In the proposed ML workflow, high-level requirements (from which the ML model is developed) are expressed in a traditional form, therefore the regular normal-range requirements-based testing approach described in DO-178C 6.4.2.1 can be applied to achieve this objective.\\
\hline
2&Executable object code is robust with high-level requirements.
&x&x&x&x&&
Same as for Objective~1, traditional robustness (abnormal range) requirements-based testing approach described in DO-178C 6.4.2.2 can be applied to achieve this objective.\\
\hline
3&Executable object code complies with low-level requirements.
&i&i&x&&&
This objective does not apply to Level~D and is excluded from the analysis.\\
\hline
4&Executable object code is robust with low-level requirements.
&i&x&x&&&
This objective does not apply to Level~D and is excluded from the analysis.\\
\hline
5&Executable object code is compatible with target computer.
&x&x&x&x&&
This objective can be achieved by execution of hardware/software integration tests in the target environment; there is no difference between ML and traditional components.\\
\hline
\end{tabular}
\end{table*}

\section{Other Applicable Standards}
\label{sec:OTHER}
\subsection{DO-331}
DO-331 \cite{RTCA.331} is a supplement to the DO-178C standard, which provides additional guidance when a model-based approach is used for software development and verification. DO-331 provides modifications and additions to DO-178C objectives and processes addressing special aspects of model-based design approach. ML technology has certain similarities with the model-based design described in DO-331:
\begin{itemize}
    \item the ML model is used to capture the intended functional behavior of a (software) system being developed, i.e., the ML model can express system or software requirements.
    \item the ML model can be executed using test data sets to demonstrate certain ML model properties (e.g., compliance with requirements, from which an ML model is developed). This is similar to the model simulation concept described in MB.4.4.4 of DO-331.
\end{itemize}
Based on these similarities, the assumption for positioning an ML model as low-level software requirements is made (DO-331, MB example 5). Moreover, ML model testing can be claimed as a means of compliance for most of the verification objectives for low-level requirements (DO-331, Item 1 in MB.A-4). In this case, some additional objectives for data set verification with respect to correctness and completeness are imposed (similarly to additional objectives for simulation cases described in MB.6.8.3.2 of DO-331). However, ML model testing does not resolve the ML-specific {\em traceability\/} and {\em coverage\/} issues.

\subsection{DO-254}
An ML-based system can be implemented using different types of electronic hardware such as microprocessor/microcontrollers, programmable logic devices (PLD), graphics processing units (GPU), or application specific integrated circuits (ASICs). Guidance materials for airborne electronic hardware development include DO-254 \cite{RTCA.254} complemented by the more recent EASA CM–SWCEH-001 \cite{EASA.CM-HW-001} and FAA Order 8110.105 \cite{FAA.8110.105}, which provide more specific guidance for modern hardware technologies such as  highly-complex commercial off-the-shelf (COTS) microprocessor/microcontrollers (e.g., with multi-core processing units) or GPUs.

For ML systems implemented as software hosted on COTS microprocessors or microcontrollers, development assurance of the core processing unit executing ML software is covered by the application of DO-178C for software, and no special hardware consideration is required with respect to ML (9.2 of \cite{EASA.CM-HW-001}). 

For ML systems implemented using PLD or ASIC, DO-254 with additional considerations provided in \cite{EASA.CM-HW-001} and \cite{FAA.8110.105} apply. The same \textit{traceability} and \textit{coverage} issues that we described above for the DO-178C objectives are relevant for the HDL language, which is typically used for PLD/ASIC design. Nevertheless, for Level~D hardware, traceability of low-level requirements is not required (DO-254, Table A-1), and HDL coverage objectives do not apply (\cite{EASA.CM-HW-001}, 8.4.2.1). So, these issues do not affect the objectives for Level~D hardware.

DO-254 does not provide specific objectives for COTS GPUs and \cite{EASA.CM-HW-001} guidance is limited to airborne display applications only. This may not be the case for ML applications, which may leverage GPUs in a broader context. However, the existing guidelines for GPUs \cite{EASA.CM-HW-001} do not introduce any additional objectives for Level~D systems. We consider that assurance of COTS GPUs can be handled in the same manner as COTS microcontrollers and microprocessor (covered by the application of DO-178C for ML software, e.g., CUDA source code). Same arguments can be applied to neuromorphic processors. It must be noted, that multi-core  architectures typical for GPUs and neuromorphic processors are a certification challenge on its own \cite{matters2018multicore}, which must be taken into account.

\subsection{System level assurance standards}
A system is defined in DO-178C as "a collection of hardware and software components organized to accomplish a specific function or set of functions". ARP4754A
\cite{SAE.4754}, the primary standard for design assurance at system level, includes objectives for system level requirements and architecture definition and validation, integration and verification of system components, as well as generic configuration management, process assurance, and certification objectives. These objectives are formulated as 'black-box' goals, abstracted from the technologies used for design, implementation, and assurance of system components and therefore achievable for ML-based components using standard methods.

\section{Conclusions and Future Work}
\label{sec:concl}
We analyzed current industry standards for development of airborne software, electronic hardware and systems with respect to low-criticality ML system of design assurance Level~D. We studied the incompatibilities between the standards and aspects of ML technology, which may hinder the fulfillment of the standards' objectives: \textit{traceability, coverage}, and \textit{ML model verification issues}. To mitigate these issues, we proposed the assumptions for an ML development workflow, limiting the scope to non-adaptive, supervised learning systems with a specific breakdown between system and software/hardware artifacts inspired by the model-based design approach. Finally, we reviewed the applicable objectives of the selected standards in the context of the proposed workflow and have shown that under the proposed assumptions all applicable objectives of DO-178C, DO-331, DO-254, and ARP-4754 can be achieved for Level~D ML-based systems using standard activities and methods.

For future work we are planning to extend the analysis to higher assurance levels. We will study new techniques that might be able to address the identified \emph{explainability} and \emph{coverage} issues. We also want to investigate how the analysis can be extended toward online learning and adaptive systems.

\
\
\noindent{\bf Acknowledgments.} We would like to thank Shanza Zafar, Kevin Schmiechen and Pranav Nagarajan for the careful review of this paper and helpful feedback.

\bibliographystyle{IEEEtran}
\bibliography{bibliography}

\begin{thebibliography}{10}
\providecommand{\url}[1]{#1}
\csname url@samestyle\endcsname
\providecommand{\newblock}{\relax}
\providecommand{\bibinfo}[2]{#2}
\providecommand{\BIBentrySTDinterwordspacing}{\spaceskip=0pt\relax}
\providecommand{\BIBentryALTinterwordstretchfactor}{4}
\providecommand{\BIBentryALTinterwordspacing}{\spaceskip=\fontdimen2\font plus
\BIBentryALTinterwordstretchfactor\fontdimen3\font minus
  \fontdimen4\font\relax}
\providecommand{\BIBforeignlanguage}[2]{{%
\expandafter\ifx\csname l@#1\endcsname\relax
\typeout{** WARNING: IEEEtran.bst: No hyphenation pattern has been}%
\typeout{** loaded for the language `#1'. Using the pattern for}%
\typeout{** the default language instead.}%
\else
\language=\csname l@#1\endcsname
\fi
#2}}
\providecommand{\BIBdecl}{\relax}
\BIBdecl

\bibitem{easaroadmap2020}
``Artificial intelligence roadmap. a human-centric approach to {AI} in
  aviation,'' European Aviation Safety Agency, Tech. Rep., 2020.

\bibitem{RTCA.178}
\emph{Software Considerations in Airborne Systems and Equipment Certification},
  RTCA, Inc. Std. RTCA DO-178C, 2011.

\bibitem{SAE.4754}
\emph{Guidelines for Development of Civil Aircraft and Systems}, SAE
  International Std. SAE ARP4754A, 2010.

\bibitem{runway-debris-detection-2018}
H.~Xu, Z.~Han, S.~Feng, H.~Zhou, and Y.~Fang, ``Foreign object debris material
  recognition based on convolutional neural networks,'' \emph{EURASIP Journal
  on Image and Video Processing}, no.~21, 2018.

\bibitem{runway-object-detection-2020}
J.~P~Sajeevan, P.~Prasanna~Kumar, S.~G~P, and S.~Mishra, ``{DNN} design for
  object detection in airport runway operations,'' \emph{International Journal
  of Advanced Science and Engineering}, vol.~6, pp. 223--225, 12 2020.

\bibitem{CNN-runway-incursions-2019}
Z.~Chen and J.~Juang, ``\BIBforeignlanguage{English}{Mitigation of runway
  incursions by using a convolutional neural network to detect and identify
  airport signs and markings},'' \emph{\BIBforeignlanguage{English}{Sensors and
  Materials}}, vol.~31, no.~12, pp. 3947--3958, 2019.

\bibitem{IFCS2007}
J.~Bosworth and P.~Williams-Hayes, ``Flight test results from the nf-15b
  intelligent flight control system ({IFCS}) project with adaptation to a
  simulated stabilator failure,'' NASA, Tech. Rep. NASA/TM-2007-214629, 2007.

\bibitem{YuYaoLiu2018}
Y.~Yu, H.~Yao, and Y.~Liu, ``Physics-based learning for aircraft dynamics
  simulation,'' in \emph{Proc. Conference of the Society for Prognostics and
  Health Management (PHM)}, 2018.

\bibitem{taxinet2020}
D.~J. Fremont, J.~Chiu, D.~D. Margineantu, D.~Osipychev, and S.~A. Seshia,
  ``Formal analysis and redesign of a neural network-based aircraft taxiing
  system with {VerifAI},'' in \emph{Computer Aided Verification}, S.~K. Lahiri
  and C.~Wang, Eds.\hskip 1em plus 0.5em minus 0.4em\relax Cham: Springer
  International Publishing, 2020, pp. 122--134.

\bibitem{vision-based-landing-2019}
Y.~Bicer, M.~Moghadam, C.~Sahin, B.~Eroğlu, and N.~Üre, ``Vision-based {UAV}
  guidance for autonomous landing with deep neural networks,'' 01 2019.

\bibitem{easaL1concept2021}
``{EASA Concept Paper: First usable guidance for Level 1 machine learning
  applications},'' European Aviation Safety Agency, Tech. Rep., 2021.

\bibitem{ACAS-XU-2018}
\BIBentryALTinterwordspacing
K.~D. Julian, M.~J. Kochenderfer, and M.~P. Owen, ``Deep neural network
  compression for aircraft collision avoidance systems,'' \emph{CoRR}, vol.
  abs/1810.04240, 2018. [Online]. Available:
  \url{http://arxiv.org/abs/1810.04240}
\BIBentrySTDinterwordspacing

\bibitem{HEO2018470}
\BIBentryALTinterwordspacing
S.~Heo and J.~H. Lee, ``Fault detection and classification using artificial
  neural networks,'' \emph{IFAC-PapersOnLine}, vol.~51, no.~18, pp. 470--475,
  2018, 10th IFAC Symposium on Advanced Control of Chemical Processes ADCHEM
  2018. [Online]. Available:
  \url{https://www.sciencedirect.com/science/article/pii/S2405896318320664}
\BIBentrySTDinterwordspacing

\bibitem{DNN-diagnostics-2019}
J.~Yang, Y.~Guo, and W.~Zhao, ``Aircraft actuator fault diagnosis using deep
  learning based sparse representation and {TSM},'' in \emph{2019 IEEE
  Aerospace Conference}, 2019, pp. 1--9.

\bibitem{aerospace8030065}
\BIBentryALTinterwordspacing
Y.~Lin, ``Spoken instruction understanding in air traffic control: Challenge,
  technique, and application,'' \emph{Aerospace}, vol.~8, no.~3, 2021.
  [Online]. Available: \url{https://www.mdpi.com/2226-4310/8/3/65}
\BIBentrySTDinterwordspacing

\bibitem{lowry2019}
M.~Lowry, T.~Pressburger, D.~A. Dahl, and M.~Dalal, ``Towards autonomous
  piloting: Communicating with air traffic control,'' in \emph{AIAA Scitech
  Forum}, 2019, p. 2207.

\bibitem{Schumann2010}
\BIBentryALTinterwordspacing
J.~Schumann, P.~Gupta, and Y.~Liu, \emph{Application of Neural Networks in High
  Assurance Systems: A Survey}.\hskip 1em plus 0.5em minus 0.4em\relax Berlin,
  Heidelberg: Springer Berlin Heidelberg, 2010, pp. 1--19. [Online]. Available:
  \url{https://doi.org/10.1007/978-3-642-10690-3_1}
\BIBentrySTDinterwordspacing

\bibitem{eurocae2021SoC}
``Artificial intelligence in aeronautical systems. statement of concerns.''
  EUROCAE, Tech. Rep. AIR6988, 2021.

\bibitem{delseny2021MLinCertifiedSys}
H.~Delseny, C.~Gabreau, A.~Gauffriau, B.~Beaudouin, L.~Ponsolle, L.~Alecu,
  H.~Bonnin, B.~Beltran, D.~Duchel, J.-B. Ginestet \emph{et~al.}, ``White paper
  machine learning in certified systems,'' \emph{arXiv preprint
  arXiv:2103.10529}, 2021.

\bibitem{ZHANG2020}
\BIBentryALTinterwordspacing
J.~Zhang and J.~Li, ``Testing and verification of neural-network-based
  safety-critical control software: A systematic literature review,''
  \emph{Information and Software Technology}, vol. 123, p. 106296, 2020.
  [Online]. Available:
  \url{https://www.sciencedirect.com/science/article/pii/S0950584920300471}
\BIBentrySTDinterwordspacing

\bibitem{Borgetal2018}
\BIBentryALTinterwordspacing
M.~Borg, C.~Englund, K.~Wnuk, B.~Dur{\'{a}}n, C.~Levandowski, S.~Gao, Y.~Tan,
  H.~Kaijser, H.~L{\"{o}}nn, and J.~T{\"{o}}rnqvist, ``Safely entering the
  deep: {A} review of verification and validation for machine learning and a
  challenge elicitation in the automotive industry,'' \emph{CoRR}, vol.
  abs/1812.05389, 2018. [Online]. Available:
  \url{http://arxiv.org/abs/1812.05389}
\BIBentrySTDinterwordspacing

\bibitem{aravantinos2019traceability}
V.~Aravantinos and F.~Diehl, ``Traceability of deep neural networks,'' 2019.

\bibitem{sun2019nnCoverage}
Y.~Sun, X.~Huang, D.~Kroening, J.~Sharp, M.~Hill, and R.~Ashmore, ``Structural
  test coverage criteria for deep neural networks,'' \emph{ACM Transactions on
  Embedded Computing Systems (TECS)}, vol.~18, no.~5s, pp. 1--23, 2019.

\bibitem{cheng2018quantitativeCov}
C.-H. Cheng, C.-H. Huang, and H.~Yasuoka, ``Quantitative projection coverage
  for testing {ML}-enabled autonomous systems,'' in \emph{International
  Symposium on Automated Technology for Verification and Analysis}.\hskip 1em
  plus 0.5em minus 0.4em\relax Springer, 2018, pp. 126--142.

\bibitem{HKvK2020}
\BIBentryALTinterwordspacing
S.~Heyer, D.~Kroezen, and E.-J.~V. Kampen, \emph{Online Adaptive Incremental
  Reinforcement Learning Flight Control for a CS-25 Class Aircraft}, 2020.
  [Online]. Available: \url{https://arc.aiaa.org/doi/abs/10.2514/6.2020-1844}
\BIBentrySTDinterwordspacing

\bibitem{RTCA.331}
\emph{Model-Based Development and Verification Supplement to DO-178C and
  DO-278A}, RTCA, Inc. Std. RTCA DO-331, 2011.

\bibitem{dmitriev2020leanMBD}
K.~Dmitriev, S.~A. Zafar, K.~Schmiechen, Y.~Lai, M.~Saleab, P.~Nagarajan,
  D.~Dollinger, M.~Hochstrasser, F.~Holzapfel, and S.~Myschik, ``{A Lean and
  Highly-automated Model-Based Software Development Process Based on
  DO-178C/DO-331},'' in \emph{2020 AIAA/IEEE 39th Digital Avionics Systems
  Conference (DASC)}.\hskip 1em plus 0.5em minus 0.4em\relax IEEE, 2020, pp.
  1--10.

\bibitem{DeepXplore}
\BIBentryALTinterwordspacing
K.~Pei, Y.~Cao, J.~Yang, and S.~Jana, ``{DeepXplore}: Automated whitebox
  testing of deep learning systems,'' in \emph{Proceedings of the 26th
  Symposium on Operating Systems Principles}, ser. SOSP '17.\hskip 1em plus
  0.5em minus 0.4em\relax New York, NY, USA: Association for Computing
  Machinery, 2017, p. 1–18. [Online]. Available:
  \url{https://doi.org/10.1145/3132747.3132785}
\BIBentrySTDinterwordspacing

\bibitem{RTCA.254}
\emph{Design Assurance Guidance for Airborne Electronic Hardware}, RTCA, Inc.
  Std. RTCA DO-254, 2000.

\bibitem{EASA.CM-HW-001}
\emph{Certification Memorandum. Development Assurance of Airborne Electronic
  Hardware}, European Aviation Safety Agency Std. CM–SWCEH-001 Issue 01
  Revision 02, 2018.

\bibitem{FAA.8110.105}
\emph{Simple and Complex Electronic Hardware Approval Guidance}, U.S. Federal
  Aviation Administration Std. ORDER 8110.105A, 2017.

\bibitem{matters2018multicore}
W.~Matters and I.~River, ``Certification of avionics applications on multi-core
  processors: Opportunities and challenges,'' 2018.

\end{thebibliography}

\end{document}